\pdfoutput=1
\documentclass[11pt]{article}

\usepackage[final]{acl}

\usepackage{times}
\usepackage{latexsym}
\usepackage{amssymb}
\usepackage{tikz}
\usepackage{amsmath}
\usepackage{amsthm} % Required for theorem-like environments
\usepackage{multirow}
\theoremstyle{definition}
\newtheorem{proposition}{Proposition}
\newtheorem{theorem}{Theorem}

\usepackage[T1]{fontenc}
\usepackage[utf8]{inputenc}
\usepackage{hyperref}
\usepackage{microtype}

\usepackage{inconsolata}

\usepackage{graphicx}
\usepackage{amsmath}
\definecolor{pastelCyan}{RGB}{173, 216, 230}
\title{A Conformal Risk Control Framework for Granular Word Assessment and Uncertainty Calibration of CLIPScore Quality Estimates}

\author{
  \textbf{Gonçalo Gomes\textsuperscript{1,2}},
  \textbf{Bruno Martins\textsuperscript{1,2}},
  \textbf{Chrysoula Zerva\textsuperscript{1,3}},
\\
  \textsuperscript{1}Instituto Superior Técnico, University of Lisbon
\\
  \textsuperscript{2}INESC-ID
\\
  \textsuperscript{3}Instituto de Telecomunicações
\\
  \small{
    {\{goncaloecgomes, chrysoula.zerva, bruno.g.martins\}@tecnico.ulisboa.pt}
  }
}

\begin{document}
\maketitle
\begin{abstract}
This study explores current limitations of learned image captioning evaluation metrics, specifically the lack of granular assessments for errors within captions, and the reliance on single-point quality estimates without considering uncertainty. To address the limitations, we propose a simple yet effective strategy for generating and calibrating distributions of CLIPScore values. Leveraging a model-agnostic conformal risk control framework, we calibrate CLIPScore values for task-specific control variables, tackling the aforementioned limitations. Experimental results demonstrate that using conformal risk control, over score distributions produced with simple methods such as input masking, can achieve competitive performance compared to more complex approaches. Our method effectively detects erroneous words, while providing formal guarantees aligned with desired risk levels. It also improves the correlation between uncertainty estimations and prediction errors, thus enhancing the overall reliability of caption evaluation metrics. \looseness-1
\end{abstract}

\section{Introduction}

Image Captioning (IC) evaluation is a crucial task in vision-and-language research, aiming to assess how accurately textual descriptions represent visual contents. Reference-free metrics such as CLIPScore~\cite{hesselclipscore, Gomes2025EvaluationOM}, which measure caption quality by computing the cosine similarity between image and text embeddings, have been shown to correlate strongly with human judgments. However, simply scoring captions is often insufficient, as these quality assessments can be hard to interpret or unreliable.

In many cases, effective evaluation requires not only an overall score of caption quality, but also the detection of specific errors within the caption. Without this granular information, the assessment can seem incomplete or less useful.
Beyond the lack of granularity, existing metrics provide IC quality assessments relying on single-point estimates, in the sense that a single value is used as an evaluation score, without incorporating any indication of confidence over the results.
This absence of uncertainty quantification can give way to misleading scores, reducing user trust. 

To address these challenges, we propose a conformal risk control framework, aiming at task-specific calibrated predictions, in conjunction with a simple yet effective strategy for generating distributions over CLIPScore predictions. This provides us with a principled way to adapt IC evaluation both to fine-grained analysis for each caption, and to a broader view of performance over a dataset, allowing for user-defined criteria to determine risk. 

First, we improve interpretability by detecting foils, which are words in the caption that are not associated with the image. Second, we overcome the limitations of single-point evaluation by introducing well-calibrated intervals, providing a trustworthy measure of caption reliability.

Experimental findings demonstrate that using conformal risk control, over the distributions produced with simple methods for expressing uncertainty, such as masking parts of the input, can achieve competitive performance on foil detection compared to more complex and specialized approaches. Conformal risk control can also provide improvements in the correlation between uncertainty estimations and prediction errors, enhancing the overall reliability of caption evaluation metrics. Furthermore, we emphasize that other existing state-of-the-art methods can also benefit from our conformal calibration framework, gaining formal guarantees over their results. The proposed methodology is model-agnostic, and our work underscores the adaptability and broad applicability of risk control, offering a compelling case for its integration into vision and language research.

\section{Related Work}
This section reviews prior work relevant to our research, organized into three key areas: uncertainty quantification, conformal risk control, and image captioning evaluation. 

\subsection{Uncertainty Quantification}

Over-confident predictions are a widespread issue in machine learning models, leading to efforts to integrate uncertainty quantification techniques. Traditionally, uncertainty estimation in neural networks has used Bayesian approaches, where weights are treated as probability distributions instead of fixed values ~\cite{mackay1992bayesian,welling2011bayesian,tran2019bayesian}. However, the high computational cost of these methods led to the adoption of alternative solutions, such as model variance approaches, to approximate uncertainty.

One common method involves deep ensembles, creating multiple neural networks and calculating the empirical variance of their output as an uncertainty measure~\cite{garmash2016ensemble,kendall2017uncertainties, pearce2020uncertainty,zhan2023test}. Another popular variance method is Monte Carlo dropout~\cite{gal2016dropout,glushkova2021uncertainty}, which leverages dropout regularization at test time, performing multiple stochastic forward passes and calculating the mean and variance of the outputs to approximate uncertainty. However, recent work by \citet{ulmer2021know} has shown that these variance-based methods can become unstable in out-of-distribution data, failing to produce accurate uncertainty estimates.

\subsection{Conformal Risk Control}

Conformal prediction and risk control are powerful frameworks for calibrating predictive models to ensure reliable decision making, giving formal guarantees tailored to the specific risks of a given domain. As mentioned by~\citet{angelopoulos2021gentle}, conformal prediction provides a model-agnostic method for creating prediction sets that are guaranteed to contain the true outcome with a user-defined probability. This assurance is achieved by focusing on coverage as the risk function, ensuring that the prediction sets have the specified coverage level across various data distributions and models. Similarly, as discussed by \citet{bates2021distribution}, the risk control framework extends the concept of conformal prediction by incorporating broader notions of risk beyond coverage. This generalization allows the calibration of predictions to minimize specific, application-relevant risks, such as the cost of errors or misclassifications in high-stakes settings (i.e., the false positive rate). While both frameworks aim to improve predictive reliability, conformal prediction is primarily concerned with probabilistic validity, whereas risk control is the general framework that emphasizes aligning predictions with context-specific risk tolerances.

\subsection{Image Captioning Evaluation}

Recently, image captioning evaluation shifted towards the use of reference-free metrics to assess captioning models. One of the pioneering metrics in this new approach is CLIPScore~\cite{hesselclipscore}, which evaluates captions without ground-truth references. Based on the Contrastive Language Image Pretraining (CLIP) model~\cite{radford2021learning}, CLIPScore calculates a modified cosine similarity between the image and caption representations under evaluation. This approach has shown a high correlation with human judgments, outperforming established reference-based metrics such as BLEU~\cite{papineni2002bleu} and CIDEr~\cite{vedantam2015cider}. CLIPScore has become a widely adopted metric for image captioning evaluation, inspiring the development of numerous new learned CLIP-based evaluation metrics~\cite{sarto2023positive,kim2022mutual,Gomes2025EvaluationOM}.

However, scoring alone is insufficient for a comprehensive evaluation, which has led to recent studies focused on identifying foil words in captions. \citet{shekhar2017foil} introduced the FOIL-it benchmark, featuring data with foil words by replacing nouns in MS-COCO~\cite{lin2014microsoft} captions with semantically similar alternatives. Building on this foundation, ALOHa~\cite{petryk2024aloha} expanded the scope by addressing foil words involving a broader range of objects, particularly visual concepts under-represented in training data for captioning models ~\cite{agrawal2019nocaps}. 

In terms of recent methods for detecting foil words, Rich-HF \cite{liang2024rich} employs human-annotated datasets of mismatched keywords and implausible image regions, to train a multimodal language model capable of providing dense alignment feedback. In turn, \citet{nam2024extract} introduced a novel approach for detecting foil words using pre-trained CLIP models. Their method refines gradient-based attribution computations, leveraging negative gradients of individual text tokens as indicators of foil words.

\begin{figure*}
    \centering
    \includegraphics[width=\linewidth]{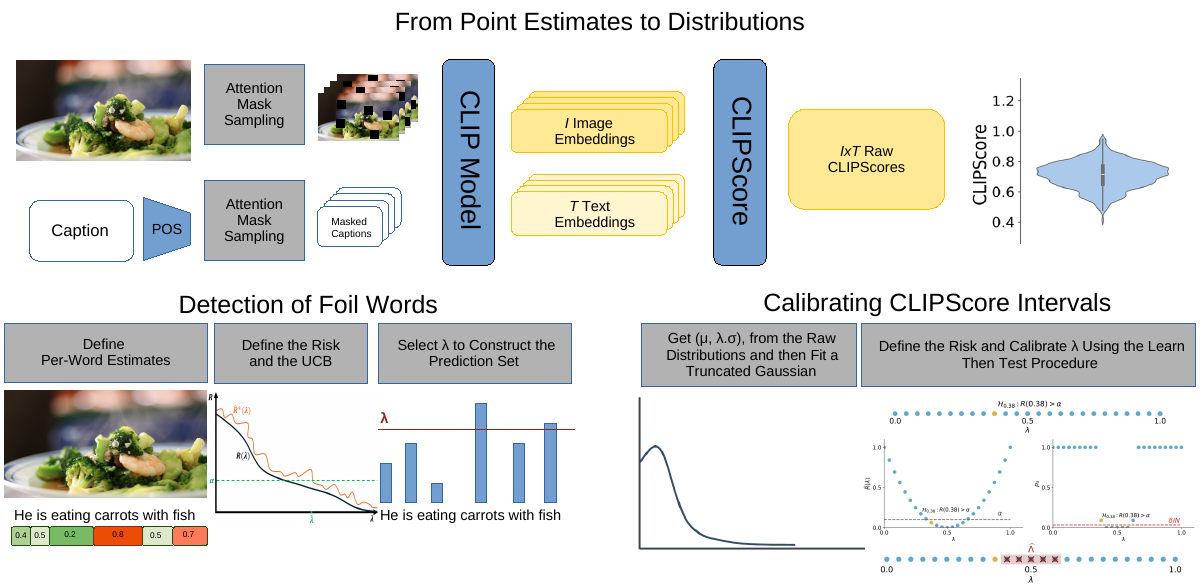}
    \caption{General overview on the proposed approach, using conformal risk control over CLIPScore values for two particular applications, namely the detection of foil words and the production of calibrated CLIPScore intervals.}
    \label{fig:Diagram}
\end{figure*}
\vspace{-0.25cm}

\section{From Point Estimates to Distributions}
\label{sec:attention_mask}

%We aim to extend the CLIPScore approach to produce a distribution of scores capturing the compatibility between a textual caption and an image, instead of a single point estimate. Appendix \ref{app:clipscore} details the computation of CLIPScore values, according to the original proposal. \chryssa{This paragraph can be skipped if we need space}

Techniques like deep ensembles or Monte Carlo (MC) dropout, commonly used to construct output distributions from instance regressor systems, are not fully model-agnostic approaches, nor are they suited to our specific objectives. Deep ensembles are unsuitable since we aim to measure the uncertainty of scores produced with individual publicly available CLIP models, without further training. In turn, MC dropout is impractical since CLIP models generally lack dropout layers.

We propose an alternative strategy for producing score distributions that express uncertainty, leveraging attention masks over the CLIP vision and text encoders to generate output distributions, by randomly masking portions of the input data. The top part from Figure \ref{fig:Diagram} illustrates this procedure. We create $I$ samples for images by randomly masking $\xi_i$\% of the image patches from the computations associated to the self-attention operations. For captions, we generate $T$ samples by randomly masking $\xi_t$\% of the tokens, corresponding to words of specific parts of speech, namely nouns, proper nouns, numerals, verbs, adjectives, and adverbs. This strategy allows us to produce $I$ image embeddings and $T$ text embeddings, which can be combined to compute $I \times T$ different CLIPScore values, following the procedure outlined in Appendix \ref{app:clipscore}. 

The CLIP model we used relies on a RoBERTa-based text encoder, which may split a word into multiple subword tokens, while our POS tagging model\footnote{\url{https://stanfordnlp.github.io/stanza/pos.html}} operates at the word level. To reconcile word-level POS tagging with subword tokenization, we first map each subword token to its corresponding word. During sampling, we randomly select the words we want to mask and then identify the subword tokens associated with each sampled word. The attention for all the corresponding tokens is then masked accordingly.

\begin{figure}
    \centering
    \includegraphics[width=\linewidth]{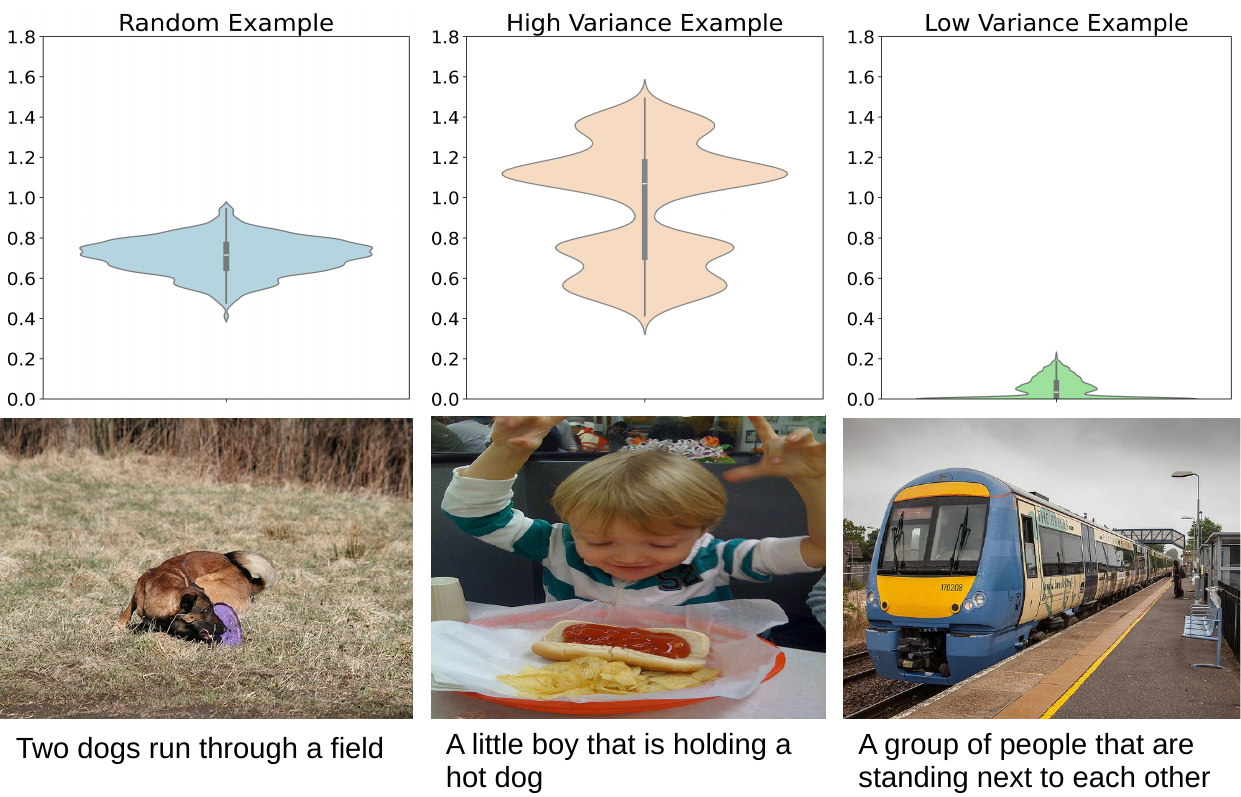}
    \caption{Violin plots with the CLIPScore distributions for three cases from the VICR dataset: a random image-caption pair, a high-variance instance identified by our method, and a low-variance instance.}
    \label{fig:violin_plots}
\end{figure}

Figure~\ref{fig:violin_plots} presents violin plots of the CLIPScore distributions produced according to our method.

\vspace{-0.25cm}

\section{Conformal Detection of Caption Errors} \label{sec:caption_errors}

In this section, we describe the application of conformal risk control for detecting foil words in image-caption pairs. The overall approach is illustrated in the bottom-left part of Figure \ref{fig:Diagram}. Leveraging the attention mask sampling method that was described in Section \ref{sec:attention_mask}, we can calibrate a control variable $\lambda$ that acts as a threshold to identify foil words in the caption. Empirical results show that this method provides good performance across several well-established benchmarks in the field~\cite{shekhar2017foil,petryk2024aloha,liang2024rich}. Furthermore, we compare the results of our simple, yet robust and well-calibrated method, against more complex, specialized, and state-of-the-art approaches, underscoring its advantages and overall effectiveness.

\subsection{Deriving Per-Word Error Estimates}
The proposed attention mask sampling method generates the output CLIPScore distribution by systematically masking parts of the input. This process inherently facilitates the evaluation of each word's contribution to the overall CLIPScore value. 

First we perform $T$ iterations of the text encoder mask sampling process. For each iteration, we mask a set of words in the caption, $W_t$, using the attention mask in the text encoder to produce a text mask embedding ($E_t^M$). For each masked word $w_j$ we keep track of its index $j$ in the original caption. We define $E_C$ as the text embedding of the original caption. Then, we compute the CLIPScore difference between the resulting text mask embedding and the original caption text embedding, with respect to $I$ image embeddings generated by randomly masking patches of the image ($E^M_i$) (see Section \ref{sec:attention_mask}).   
The degree of contribution of $W_t$ to the original CLIPScore can be quantified as the average of this difference over the $I$ images, as formally described in Equation  \ref{eq:error_score}. \looseness=-1
\begin{equation}\label{eq:error_score} 
    \resizebox{.88 \columnwidth}{!}{$
    v_t = \frac{1}{I} \sum\limits_{i=1}^{I} \left(\text{CLIPS}(E_t^M, E^M_i)- \text{CLIPS}(E_C, E^M_i) \right).
    $}
\end{equation}

Note that a positive difference indicates that the masked words negatively contributed to the CLIPScore value in the original caption. Consequently, these words are more likely to act as foil words, which diminish the overall relevance or coherence of the caption in relation to the image.

Next, we aggregate the results of Equation 1 over the indexes $j$ of the masked words, obtaining the average error scores $V[j]$, as follows:
\begin{equation}\label{eq:cum_error_scores} 
    V[j] = \frac{1}{\sum_{t}^T \mathbf{1}_{\{w_j \in W_t\}}} \sum_t^T v_t \cdot \mathbf{1}_{\{w_j \in W_t\}}.
\end{equation}

To create the error score vector $f_v$, we apply a sigmoid transformation, $\sigma(\cdot)$, to $V$, such that
\begin{equation}\label{eq:f_s} 
    f_v[j] = \sigma(V[j]).
\end{equation}

While the application of the sigmoid function does not enhance performance, it confines the error scores to a finite range, facilitating the implementation of the conformal risk control framework.

\subsection{Risk Control on Word Error Detection}

The previous method can already help identifying the most likely inadequate word, as the one with the highest score in $f_v$ from Equation \ref{eq:f_s}. However, simply taking the word with the highest score falls short in two scenarios: cases where we consider at most one word per caption (or none) to be incorrect, which can be seen as multi-class problems,
%(i.e., the different words, plus one extra label for none, are seen as a set of classes from which one should be picked)
and cases where multiple words can be incorrect, which can be seen as multi-label problems. To address the limitations, we introduce a threshold-based approach to determine which words should be classified as errors. Specifically, we aim to obtain prediction sets $\mathcal{S}_\lambda(x)$ of foil words, as follows:
\begin{equation}\label{eq:prediction_set}
    \mathcal{S}_\lambda(x) = \{x : f_v(x) > \lambda\},
\end{equation}
where the control variable $\lambda$ acts as a threshold.

Ideally, we aim to optimize the selection of $\lambda$ so that our prediction sets meet specific user requirements regarding caption quality and error detection. For example, in some tasks, we may prioritize minimizing the false positive rate to ensure that only highly reliable foil words are included, while in others, we may focus on reducing the false negative rate to avoid missing potentially important errors. The choice of $\lambda$ can alternatively be calibrated to strike the right balance between precision and recall, depending on the task's objectives. To be able to account for these requirements, we rely on conformal risk control \cite{angelopoulos2022conformal}, since it allows control over different performance criteria, providing statistical guarantees on their bounds. Specifically, let us assume $ R(\lambda) $ is a non-increasing and monotonic function of $ \lambda $, corresponding to our preferred quality criteria. This function serves as a performance metric for $ \mathcal{S}_\lambda $, offering an interpretable assessment of its quality.

We can then use a calibration set to get the optimal parameter $ \hat{\lambda} $ while ensuring formal guarantees about the risk level. Specifically, for a user-defined risk tolerance $ \alpha $ and error rate $ \delta $, we aim to satisfy:
\begin{equation}\label{eq:risk_control}
    \mathbb{P} \big(R(\hat{\lambda}) < \alpha \big) \geq 1 - \delta.
\end{equation}

The procedure that we use to find $ \hat{\lambda} $, in order to satisfy the Inequality \ref{eq:risk_control}, assumes that we have access to a pointwise Upper Confidence Bound (UCB) for the risk function for each value of $\lambda$:
\begin{equation}\label{eq:ucb}
    \mathbb{P} \big(R(\lambda) \leq \underbrace{\hat{R}^+(\lambda)}_{\text{UCB}} \big) \geq 1 - \delta.
\end{equation}
We can then choose $\hat{\lambda}$ as the smallest value of $\lambda$ such that the entire confidence region to the right of $\hat{\lambda}$ falls below the target risk tolerance $\alpha$:
\begin{equation}\label{eq:threshold}
    \hat{\lambda} = \text{inf} \thinspace \Big\{ \lambda \in \Lambda: \hat{R}^+(\lambda) \leq \alpha, \thinspace \forall\lambda' \geq \lambda \Big\}.
\end{equation}

%Figure \ref{fig:risk_control} contains a visualization of the upper bound calibration process.

%\begin{figure}
%    \centering
%    \includegraphics[width=1\linewidth]{Figures/risk_control.png}
%    \caption{Visualization of upper confidence bound calibration}
%    \label{fig:risk_control}
%\end{figure}

As mentioned by \citet{bates2021distribution}, the bound guarantees, that act as foundations to obtain the conformal risk-controlling prediction sets, work as long as we have access to a concentration result. In other words, they work as long as we have a mathematical guarantee that the risk is tightly bounded (i.e., controlled), and does not deviate too much from its expected value. Therefore, we can construct the UCB for the risk using concentration inequalities. This approach leverages the empirical risk, which is computed by averaging the loss of the set-valued predictor $\mathcal{S}_\lambda$ over a calibration set. The empirical risk is defined as:
\begin{equation}\label{eq:empirical_risk}
    \hat{R}(\lambda) = \frac{1}{n}\sum_{i=1}^n \mathcal{L}(Y_i, \mathcal{S}_\lambda(X_i)),
\end{equation}
where $n$ is the size of the calibration set, $\mathcal{L}(Y_i, \mathcal{S}_\lambda(X_i))$ represents the loss for each pair $(Y_i, X_i)$, and $\mathcal{S}_\lambda(X_i)$ is the prediction generated by the set-valued predictor for input $X_i$.

A concentration inequality provides bounds on the tail probabilities of a random variable, and it is typically expressed in the following form:
\begin{equation}
    \mathbb{P}\Big(|\hat{R}(\lambda) - R(\lambda)| \geq \epsilon \Big) \leq h(\epsilon;R(\lambda)),
\end{equation}
where $h(\epsilon;R(\lambda))$ is a non-increasing function of $\epsilon > 0$ and depends on the parameter $R(\lambda)$. By appropriately rearranging this inequality, we can control either the lower or upper tail probability.

In general, a UCB can be obtained if the lower tail probability for $\hat{R}(\lambda)$ of the concentration inequality can be controlled in the following sense:
\begin{proposition} \label{prop:ucb}
    Suppose $g(t;R)$ is a non-decreasing function in $t \in \mathbb{R}$ for every $R$:
    \begin{equation}
        \mathbb{P}\Big(\hat{R}(\lambda) \leq t \Big) \leq g(t;R(\lambda)).
    \end{equation}
    Then, $\hat{R}^+(\lambda) = \text{sup }\Big\{ R: g(\hat{R}(\lambda);R) \geq \delta \Big\}$ satisfies the Inequality \ref{eq:ucb}. The proof of Proposition \ref{prop:ucb} can be found in Appendix \ref{appendix:proof_prop_1}.
\end{proposition}

There are numerous concentration inequalities to choose from. In this work, we opted for a combination of Hoeffding and Bentkus bounds~\cite{bentkus}\footnote{Exploring other alternatives could lead to the discovery of even tighter bounds for this use case, but this option was considered out of scope for this work.}.
We can obtain a tighter lower tail probability bound for $\hat{R}(\lambda)$, combining Propositions \ref{prop:hoeffding} and \ref{prop:bentkus}, described in Appendix \ref{app:concentrations}. We thus have
\begin{equation*}
\resizebox{\hsize}{!}{%
    $g^{HB}(t;R(\lambda)) = \text{min}\Big( g^{H}(t;R(\lambda)), g^{B}(t;R(\lambda))\Big),$ 
    }
\end{equation*}
where $g^{H}(t;R(\lambda))$ and $g^{B}(t;R(\lambda))$ refer to the Hoeffding and Bentkus lower tail probability bounds, respectfully.

Applying Proposition \ref{prop:ucb}, we obtain a $(1-\delta)$ upper confidence bound for $R(\lambda)$ as:
\begin{equation*}
    \hat{R}_{HB}^+(\lambda) = \text{sup }\Big\{ R: g^{HB}(\hat{R}(\lambda);R) \geq \delta \Big\}.
\end{equation*}

We can now determine the optimal threshold $\hat{\lambda}$ for calibrating the prediction sets $\mathcal{S}_\lambda(x)$, as defined in Equation \ref{eq:prediction_set}, by using the upper bound risk from $\hat{R}_{HB}^+(\lambda)$ and applying it in Equation \ref{eq:threshold}. This selection for the control variable ensures a formal guarantee that the user-defined risk remains controlled within the specified tolerance, as described in Equation \ref{eq:risk_control}, even if the test data deviates slightly from the calibration distribution. However, this guarantee holds only as long as the distribution shift is not too severe, preserving the validity of the concentration result assumption.

\subsection{Experimental Results}

This section presents the datasets, the evaluation metrics, and the results for foil word recognition using the proposed method. For all experiments, we apply our methods on the multilingual LAION ViT-B/32 and ViT-H/14 CLIP models, as they have shown robust performance on English data \cite{schuhmann2022laion, Gomes2025EvaluationOM}.

\subsubsection{Datasets and Evaluation Metrics}
To ensure a fair and comprehensive evaluation, we used three well-established test benchmarks:
\vspace{0.25cm}

\textbf{FOIL-it}: $198,960$ pairs~\cite{shekhar2017foil};

\textbf{FOIL-nocaps}: $5,000$ pairs~\cite{petryk2024aloha};

\textbf{Rich-HF}: $955$ pairs~\cite{liang2024rich}.

The three datasets, which are further detailed in Appendix~\ref{app:datasets}, associate images with either correct captions or captions containing intentional errors. Among them, FOIL-it and FOIL-nocaps are constructed using the same underlying methodology: one object is replaced by a conceptually similar word (i.e., \textit{dog} can be replaced by \textit{cat}). FOIL-nocaps, built on the nocaps dataset~\cite{agrawal2019nocaps}, includes a broader range of visual concepts not typically found in standard training or evaluation datasets, which are often limited to the object classes defined in MS-COCO~\cite{lin2014microsoft}. It combines in-domain and out-of-domain captions, with the latter containing novel-class words that captioning models are unlikely to encounter in conventional evaluation datasets, testing our method's ability to generalize beyond familiar concepts.

Since the aforementioned datasets are word-level multi-class benchmarks, primarily focused on objects, the errors are restricted to nouns. We used the Rich-HF dataset to broaden our evaluation, considering multi-label scenarios and a more diverse range of word-level errors. This dataset comprises both AI-generated and human-written prompts resembling captions, collected from the Pick-a-Pic dataset~\cite{kirstain2023pick}. The creators of Rich-HF carefully selected photo-realistic images for their broader applicability, while ensuring a balanced representation across image categories.

Based on these three datasets, we conduct two types of assessments across two different classification tasks: a multi-class task and a multi-label task for detecting foil words in captions. The assessments are as follows:  

\textbf{Caption Classification} – Determining whether a caption is wrong. We evaluate this task using Average Precision (AP) and instance-level F1 score. 

\textbf{Word Error Detection} – Identifying specific foil words within a caption. For multi-class benchmarks we measure Location Accuracy (LA), while for multi-label tasks we use word-level precision, recall, and F1 scores.  

To calibrate the threshold in Equation~\ref{eq:threshold}, we first must define the risk function. Our goal is to detect foil words without resorting to trivial solutions of over-detecting most words as foils. To achieve this, we control the False Discovery Rate (FDR) for multi-class tasks, and the False Positive Rate (FPR) for multilabel scenarios. In Appendix \ref{app:metrics}, a more detailed explanation of each metric is provided. Those metrics serve as the target risk, enabling us to effectively evaluate the performance of the prediction sets $\mathcal{S}_\lambda$ in Equation~\ref{eq:threshold}.

\subsubsection{Assessing Multi-Class Guarantees}

\begin{table}[t!]
\centering
\renewcommand{\arraystretch}{1.3}
\resizebox{\columnwidth}{!}{%
\begin{tabular}{lccccccccc}
\multicolumn{1}{c}{} & \multicolumn{6}{c}{\textbf{All Instances}}                                            &  & \multicolumn{2}{c}{\textbf{Foil Only}} \\ \cline{2-7} \cline{9-10} 
\multicolumn{1}{c}{}                                & \multicolumn{2}{c}{\textbf{Calib. Set}} &  & \multicolumn{3}{c}{\textbf{Test Set}}    &  & \multicolumn{2}{c}{\textbf{Test Set}}  \\ \cline{2-3} \cline{5-7} \cline{9-10} 
\multicolumn{1}{c}{\textbf{$\alpha$}}                                & \textbf{FDR}        & \textbf{F1}       &  & \textbf{FDR} & \textbf{AP} & \textbf{F1} &  & \textbf{LA}     & \textbf{$\textbf{LA}_{\textbf{Set}}$}     \\ \hline
%\textbf{5\%}                                        & $4,77$              & $58,17$           &  & $5,05$       & $60,33$     & $58,25$     &  & $27,39$         & $27,68$              \\
\textbf{10\%}                                       & $9,69$              & $61,74$           &  & $10,10$      & $60,75$     & $61,93$     &  & $33,68$         & $34,39$              \\
\textbf{15\%}                                       & $14,62$             & $63,12$           &  & $15,02$      & $60,34$     & $63,31$     &  & $37,33$         & $38,53$              \\
\rowcolor{pastelCyan} \textbf{20\%}                                       & $19,58$             & $63,55$           &  & $20,20$      & $59,68$     & $63,76$     &  & $40,15$         & $41,92$              \\
\textbf{25\%}                                       & $24,55$             & $63,21$           &  & $25,13$      & $58,92$     & $63,56$     &  & $42,33$         & $44,69$              \\
\textbf{30\%}                                       & $29,52$             & $62,77$           &  & $30,24$      & $58,04$     & $62,81$     &  & $44,07$         & $47,06$              \\
\textbf{35\%}                                       & $34,50$             & $61,90$           &  & $35,25$      & $57,24$     & $61,81$     &  & $45,60$         & $49,31$              \\
\textbf{40\%}                                       & $39,49$             & $60,65$           &  & $40,16$      & $56,44$     & $60,49$     &  & $46,82$         & $51,18$              \\
\textbf{45\%}                                       & $44,48$             & $58,86$           &  & $45,11$      & $55,58$     & $58,76$     &  & $47,88$         & $53,11$              \\
\textbf{50\%}                                       & $49,47$             & $56,72$           &  & $50,27$      & $54,71$     & $56,68$     &  & $48,81$         & $54,88$              \\ \hline
\end{tabular}
}
\vspace{-0.25cm}
\caption{Calibration results for risk control using the multilingual LAION ViT-B/32 CLIP model, with the FOIL-it dataset as the calibration and test set. The highlighted row corresponds to the best calibration F1 score.}
\label{tab:foilit-results}
\vspace{-0.25cm}
\end{table}

\begin{table*}[t!]
\centering
\renewcommand{\arraystretch}{1.15}
\resizebox{2\columnwidth}{!}{%
\begin{tabular}{lccccccccccccccccccc}
 & \multicolumn{3}{c}{} & \multicolumn{16}{c}{\textbf{FOIL-nocaps}} \\ \cline{5-20} 
& \multicolumn{3}{c}{\textbf{FOIL-it}} & \textbf{} & \multicolumn{3}{c}{\textbf{Overall}}     & \textbf{} & \multicolumn{3}{c}{\textbf{In Domain}}   & \textbf{} & \multicolumn{3}{c}{\textbf{Near Domain}} & \textbf{} & \multicolumn{3}{c}{\textbf{Out of Domain}}  \\ \cline{2-4} \cline{6-8} \cline{10-12} \cline{14-16} \cline{18-20} 
\textbf{Model} & \textbf{FDR}      & \textbf{AP}     & \textbf{LA}     & \textbf{} & \textbf{FDR} & \textbf{AP} & \textbf{LA} & \textbf{} & \textbf{FDR} & \textbf{AP} & \textbf{LA} & \textbf{} & \textbf{FDR} & \textbf{AP} & \textbf{LA} & \textbf{} & \textbf{FDR} & \textbf{AP} & \textbf{LA} \\ \hline
\textbf{CHAIR~\cite{rohrbach2018object}}                  & $-$                 & $92,5$            & $79$              &           & $-$            & $58,3$        & $14,4$        &           & $-$            & $57,8$        & $13,5$        &           & $-$            & $59,1$        & $17,6$        &           & $-$            & $58,1$        & $12,2$        \\
\textbf{Aloha~\cite{petryk2024aloha}}                  & $-$                 & $61,4$            & $40$              &           & $-$            & $69,5$        & $45,2$        &           & $-$            & $71,8$        & $47,4$        &           & $-$            & $66,7$        & $47,3$        &           & $-$            & $70,9$        & $48,8$        \\
\textbf{GAE\_B~\cite{nam2024extract}}                 & $-$                 & $71,4$            & $73,2$            &           & $-$            & $69,0$        & $60,3$        &           & $-$            & $67,3$        & $54,7$        &           & $-$            & $68,4$        & $59,7$        &           & $-$            & $71,3$        & $63,2$        \\
\textbf{GAE\_H~\cite{nam2024extract}}                 & $-$                 & $80,6$            & $83,6$            &           & $-$            & $79,4$        & $71,6$        &           & $-$            & $78,9$        & $66,1$        &           & $-$            & $79,3$        & $70,8$        &           & $-$            & $80,2$        & $74,8$        \\ \hline
\textbf{Our Method with ML LAION ViT-B/32}                 & $20,2$              & $59,7$            & $40,2$            &           & $18,6$         & $64,4$        & $54,9$        &           & $20,4$         & $70,0$        & $53,5$        &           & $19,6$         & $72,2$        & $56,3$        &           & $16,2$         & $74,4$        & $52,6$        \\
\textbf{Our Method with ML LAION ViT-H/14}               &    $19,8$               &     $63,4$            &    $51,4$             &           & $19,1$         & $65,7$       & $60,3$        &           & $19,2$         & $70,4$        & $56,7$        &           & $19,4$         & $72,5$        & $63,0$        &           & $18,5$         & $74,0$        & $56,2$        \\ \hline
\end{tabular}
}
\vspace{-0.25cm}
\caption{Results for the calibrated sampling method %using the multilingual LAION ViT-B/32 and Vit-H/14 models 
on the FOIL-it and FOIL-nocaps benchmarks.}
\label{tab:foil-results}
\vspace{-0.25cm}
\end{table*}

To assess conformal guarantees on the word level multi-class task, we calibrate the threshold $\lambda$ using 10\% of the FOIL-it validation set, and evaluate performance on the FOIL-it and FOIL-nocaps benchmarks. Table \ref{tab:foilit-results} presents results for different risk tolerance levels. The findings show that the proposed inequality bounds are able to efficiently align the user-defined tolerance with the observed values for the chosen quality metric (i.e., the FDR), which are consistently below but close to the chosen $\alpha$.

Increasing the risk tolerance level makes the method more permissive, classifying more words as errors. This improves word-level accuracy but reduces instance-level average precision, as more instances are classified as foils. To balance the trade-off between instance-level precision and recall, we rely on the best F1 score, on the calibration set, to select a proper risk tolerance, thus selecting $\alpha =20\%$. We then use the calibrated outputs at the selected $\alpha$ to compare against state-of-the-art methods for both FOIL-it and the more challenging FOIL-nocaps benchmarks in Table \ref{tab:foil-results}. Note that for this table, we only calibrate on FOIL-it (but not FOIL-nocaps) data. By evaluating on benchmarks with different data distributions, we can also assess the validity of the concentration result assumption.

Indeed, empirical results on the FOIL-nocaps dataset indicate a more conservative estimation, as there is a slight deviation between the controlled metric (i.e., the FDR) and the desired tolerance (Table \ref{tab:foil-results}). We attribute this to distribution differences between the calibration and test sets. Nevertheless, our method successfully controls the risk, suggesting that the distribution shift is not too severe, and that the concentration result assumption remains valid.
Additionally, %Table \ref{tab:foil-results} evaluates our calibrated method alongside current state-of-the-art systems. 
our approach achieves performance comparable to ALOHa \cite{petryk2024aloha} on the FOIL-it benchmark, and to both ALOHa and CHAIR \cite{rohrbach2018object} on FOIL-nocaps. Notably, both CHAIR and ALOHa are more complex methods, with ALOHa leveraging large language models to detect erroneous words.  
    
Although our method falls short compared to the recent approach by \citet{nam2024extract}, which employs a sophisticated gradient-based attribution technique where the negative gradient of individual text tokens signals foil words, we emphasize the simplicity of our attention sampling method to produce CLIPScore distributions, and the model-agnostic nature of our calibration framework. Unlike these more complex approaches that rely on specific architectures or gradient-based computations, our method can be applied to a wide range of models, including the current state-of-the-art systems, for further calibration to user-requirements and formal guarantee assessments. %Despite its simplicity, our approach performs on par with more complex systems while providing formal guarantees regarding risk control.
Appendix \ref{app:qualitative} provides additional qualitative analyses for the FOIL-it and FOIL-nocaps benchmarks.

\begin{table}[t!]
\centering
\renewcommand{\arraystretch}{1.3}
\resizebox{\columnwidth}{!}{%
\begin{tabular}{lcccccccccc}
\multicolumn{1}{c}{} & \multicolumn{2}{c}{\textbf{Calib. Set}} &  & \multicolumn{7}{c}{\textbf{Test Set}}                                                    \\ \cline{2-3} \cline{5-11} 
\multicolumn{1}{c}{\textbf{$\alpha$}}                                & \textbf{FPR}        & \textbf{F1}       &  & \textbf{FPR} & \textbf{AP} & \textbf{F1} &  & \textbf{PREC} & \textbf{REC} & \textbf{F1} \\ \hline
%\textbf{5\%}                                        & $3,63$              & $48,41$           &  & $3,20$       & $76,07$     & $42,15$     &  & $13,62$       & $34,63$      & $19,55$     \\
\textbf{10\%}                                       & $8,09$              & $56,87$           &  & $7,13$       & $78,22$     & $52,95$     &  & $21,03$       & $39,29$      & $27,40$     \\
\textbf{15\%}                                       & $12,65$             & $59,68$           &  & $10,74$      & $79,29$     & $58,97$     &  & $26,14$       & $43,76$      & $32,73$     \\
\rowcolor{pastelCyan} \textbf{20\%}                                       & $17,41$             & $61,40$           &  & $16,92$      & $80,73$     & $65,42$     &  & $30,43$       & $50,49$      & $37,97$     \\
\textbf{25\%}                                       & $22,16$             & $61,24$           &  & $24,03$      & $80,44$     & $66,02$     &  & $31,02$       & $56,80$      & $40,12$     \\
\textbf{30\%}                                       & $27,02$             & $58,75$           &  & $31,42$      & $80,41$     & $66,76$     &  & $31,22$       & $62,55$      & $41,65$     \\
\textbf{35\%}                                       & $31,85$             & $56,66$           &  & $36,84$      & $80,06$     & $66,40$     &  & $31,15$       & $66,47$      & $42,42$     \\
\textbf{40\%}                                       & $36,74$             & $56,30$           &  & $40,04$      & $79,48$     & $65,28$     &  & $31,76$       & $69,41$      & $43,58$     \\
\textbf{45\%}                                       & $41,75$             & $55,07$           &  & $45,02$      & $78,95$     & $64,15$     &  & $31,84$       & $72,86$      & $44,32$     \\
\textbf{50\%}                                       & $46,67$             & $54,17$           &  & $48,64$      & $78,25$     & $62,37$     &  & $31,80$       & $76,00$      & $44,84$     \\ \hline
\end{tabular}
}
\vspace{-0.25cm}
\caption{Risk control results with the multilingual LAION ViT-B/32 CLIP model, using the Rich-HF validation set for calibration and the corresponding test set for evaluation. The highlighted row
corresponds to the best calibration F1 score.}
\label{tab:rich-calib}
\vspace{-0.25cm}
\end{table}

\subsubsection{Assessing Multi-Label Guarantees}

\begin{table}[t!]
\centering
\renewcommand{\arraystretch}{1.3}
\resizebox{\columnwidth}{!}{%
\begin{tabular}{lcccc}
\textbf{Model}        & \textbf{ft.} & \textbf{PREC} & \textbf{REC} & \textbf{F1} \\ \hline
\textbf{ALOHa ~\cite{petryk2024aloha}}        &              & $34,4$          & $31,1$         & $38,5$        \\
\textbf{Rich-HF (MH)~\cite{liang2024rich}} & \checkmark            & $43,3$          & $62,9$         & $33,0$        \\
\textbf{Rich-HF (AP)~\cite{liang2024rich}} & \checkmark            & $43,9$          & $61,3$         & $34,1$        \\
\textbf{GAE\_B~\cite{nam2024extract}}       &              & $39,8$          & $32,8$         & $50,4$        \\
\textbf{GAE\_H~\cite{nam2024extract}}       &              & $42,7$          & $36,5$         & $51,6$        \\ \hline
\textbf{Our Method with ML LAION ViT-B/32}      &              & $31,2$          & $62,6$        & $41,7$       \\
\textbf{Our Method with ML LAION ViT-H/14}        &             &    $32.0$         &  $64.2$               &    $42.7$        \\ \hline
\end{tabular}
}
\vspace{-0.25cm}
\caption{Results for the calibrated sampling method %using the multilingual LAION ViT-B/32 and ViT-H/14 CLIP models 
on the Rich-HF benchmark.}
\label{tab:rich-results}
\vspace{-0.25cm}
\end{table}
To evaluate conformal guarantees in the word-level multi-label task, we calibrate our system on the validation set of Rich-HF, and assess its performance on the corresponding test set.

Table \ref{tab:rich-calib} presents results over increasing risk tolerance levels. Similarly to the multi-class results, we consistently control the risk to align with the target tolerance level in the calibration set. However, a notable discrepancy emerges between the tolerance level and the risk metric (i.e., the false positive rate) on the calibration set. This discrepancy arises primarily due to the limited size of the Rich-HF calibration set, which contains only $955$ samples. The small sample size increases the margin of error for the upper confidence bound, which is an intentional overestimation in order to achieve more general and robust guarantees of risk control, leading to more conservative threshold estimates. 

Variability in caption characteristics further affects the applicability of the thresholds. For instance, calibrating on datasets with longer captions but testing on shorter ones will lead to higher thresholds, giving rise to an undesired strict behaviour when classifying foil words. In turn, the reverse scenario, i.e., calibrating on shorter captions and testing on longer ones, can produce overly lenient thresholds. Together, these factors influence the ability to reliably control risk across diverse scenarios. Appendix \ref{app:datasets} presents a visualization highlighting the differences between the calibration and test sets of Rich-HF, supporting a better understanding of these differences.

Table \ref{tab:rich-results} compares our calibrated method with current state-of-the-art systems. Despite its simplicity and general-purpose design, our method outperforms both the LLM-based ALOHa approach, and the specialized fine-tuned model used in the Rich-HF benchmark, achieving superior F1 performance. Similarly to the multi-class experiments, our simple method achieved lower F1 scores than the more complex and recent approach by \citet{nam2024extract}, although in this case we achieved significantly higher recall. %Despite having higher recall performance, similarly to multi-class experiments, our simple method achieved lower F1 scores than the more complex and recent approach by \citet{nam2024extract}. 

% \subsection{Overall Results}
% \input{sections/sub_sections/overall_foil}

\section{Conformalized Intervals for CLIPScore}

We now test a second application of risk control over CLIPScore values, to address the limitations of single-point evaluation metrics in Image Captioning (IC) assessments, aiming to get reliable and interpretable confidence intervals for each IC score. Leveraging the uncertainty quantification method described in Section \ref{sec:attention_mask}, we fit a truncated Gaussian distribution to construct intervals. These intervals help quantify model uncertainty more effectively, providing a nuanced and trustworthy assessment of caption quality. The overall approach is illustrated in the bottom-right part of Figure \ref{fig:Diagram}.

The choice of using truncated Gaussian distributions is motivated by CLIPScore being inherently bounded, as the metric is defined as a modified cosine similarity. In addition, it allows us to define a more meaningful rescaling of initially estimated uncertainties, effectively reordering confidence intervals to align with the deviation from ground truth, as described in the following sections.

\subsection{Risk Control on Human Correlation} \label{subsec:risk_control_ci}
Calibrating confidence intervals for CLIPScore assessments is particularly challenging because CLIPScore was not trained to predict human judgment scores, but rather to correlate with them. As a result, we cannot rely on typical risk functions such as coverage \cite{zerva-martins-2024-conformalizing}, which measures the proportion of times the ground truth falls within the computed confidence intervals. A suitable risk function must account for this indirect relationship, ensuring meaningful calibration. \looseness=-1

We propose a new risk function to calibrate our intervals that does not depend on the match of scale between the output distributions and the ground truth, specifically defined as follows: 
\begin{equation} \label{eq:UPR}
    R(\lambda) = 1 - \text{ReLU}( r(|\hat{\mu}(\lambda) - y|, \hat{\sigma}(\lambda)).
\end{equation}

This risk function leverages the Uncertainty Pearson Score (UPS), denoted as: \looseness-1
\begin{equation} \label{eq:ups}
    \text{UPS} =r(|\hat{\mu}(\lambda) - y|, \hat{\sigma}(\lambda)),
\end{equation}
where $r$ is the Pearson correlation coefficient and $y$ the ground truth (human score)~\cite{glushkova2021uncertainty}.
This metric quantifies the correlation between prediction errors and uncertainty estimates. The values $\hat{\mu}(\lambda)$ and $\hat{\sigma}(\lambda)$ are derived by fitting a truncated Gaussian distribution, using the original mean $\mu$, and scaled standard deviation $\lambda \sigma$. The values for $\mu$ and $\sigma$ are obtained empirically from the CLIPScore distribution obtained via masking.

Notably, the risk function is not monotonically non-increasing. The direct application of the framework described in Section \ref{sec:caption_errors} involves the assumption of monotonicity of the risk function, otherwise we cannot extend the pointwise convergence result, from Equation \ref{eq:threshold}, into a result on the validity of a data-driven choice of $\lambda$. To address this, we propose a strategy based on the Learn Then Test (LTT) technique \cite{angelopoulos2021gentle}, which leverages the duality between tail probability bounds in concentration inequalities and conservative $p$-values. This approach enables us to identify $\hat{\lambda}$ that satisfies Equation \ref{eq:risk_control}, extending the concentration result assumption to more general and complex risks. The procedure outputs a subset $\hat{\Lambda} \subseteq \Lambda$, ensuring all selected sets $\hat{\Lambda}$ of $\lambda$ values control the user-defined risk. We describe the process below.

\textbf{Step 1:} We first define the risk tolerance $\alpha$. Our objective is to calibrate $\lambda$ such that the resulting risk level is lower than the initial one. Looking at Equation \ref{eq:UPR}, this implies maximizing a positive correlation between estimated uncertainties and deviation from the ground truth which, naturally leads to more reliable and interpretable uncertainties. Thus, we set $\alpha$ as the risk $R(\lambda)$ at $\lambda = 1$:
\begin{equation}
\alpha = 1 - \text{ReLU}( r(|\hat{\mu}(1) - y|, \hat{\sigma}(1)).
\end{equation}

\textbf{Step 2:} For each $\lambda \in \Lambda$, in which $\Lambda$ refers to the set of acceptable values, we associate the null hypothesis $\mathcal{H}_\lambda : R(\lambda) > \alpha$. Note that rejecting $\mathcal{H}_\lambda$ means the selection of a value for $\lambda$ that controls the user-defined risk.

\textbf{Step 3:} As noted by \citet{bates2021distribution}, the upper bound $g(\hat{R}(\lambda); R)$, derived from Proposition \ref{prop:ucb}, can be interpreted as a conservative $p$-value for testing the one-sided null hypothesis $ \mathcal{H}_0 : R(\lambda) > R $. Therefore, for each null hypothesis $ \mathcal{H}_\lambda $, we can compute conservative $p$-values $p_\lambda$ using $g(\hat{R}(\lambda); \alpha)$ to test the hypothesis $ \mathcal{H}_\lambda : R(\lambda) > \alpha $.

\textbf{Step 4:} Return $\hat{\Lambda} = \mathcal{A}(\{p_\lambda\}_{\lambda \in \Lambda})$, where $\mathcal{A}$ is an algorithm designed to control the Family-Wise Error Rate (FWER). This is important because, when conducting multiple hypothesis tests, the probability of making at least one Type I error increases as the number of tests grows. Each individual test has a small chance of being a false positive (e.g., $p_\lambda < 0.05$), but as more tests are performed, these small probabilities accumulate, raising the overall risk of an error.
For the case where $\Lambda = \{ \lambda : p_\lambda < \delta \}$, the FWER is given by:
\begin{equation}
    \text{FWER}(\Lambda) = 1 - (1 - \delta)^{|\Lambda|}.
\end{equation}
We will use throughout the experiments the Bonferroni correction, which tests each hypothesis at level $\delta/|\Lambda|$, ensuring that the probability of at least one failed test is no greater than $\delta$ by the union bound.
\begin{equation}
    \hat{\Lambda} = \left\{\lambda: p_\lambda < \frac{\delta}{|\Lambda|} \right\}.
\end{equation}

\textbf{Step 5:} With the set $\hat{\Lambda}$ containing all the $\lambda$ values that successfully control the user-defined risk with statistical significance, we can further refine the selection using other specific metrics on the calibration set. In this case, we aim to identify the value $\hat{\lambda}$ which maximizes the UPS. Given our chosen risk (Equation \ref{eq:UPR}), this corresponds naturally to the $\lambda$ value with the lowest $p$-value.

\subsection{Experimental Results}
This section presents the datasets, the evaluation metrics, and the results for conformalizing CLIPScore intervals using the proposed method.

\subsubsection{Datasets and Evaluation Metrics}

To ensure a fair and comprehensive evaluation, we used four well-established datasets designed to evaluate the correlation between vision-and-language model outputs and human judgments:
\vspace{0.25cm}

\textbf{VICR}: $3,161$ instances \cite{narins2024validated};

\textbf{Polaris}: $8,726$ instances \cite{narins2024validated};

\textbf{Ex-8k}: $5,664$ instances~\cite{hodosh2013framing};

\textbf{COM}: $13,146$ instances~\cite{aditya2015images}.
\vspace{0.25cm}

We will use the validation set of VICR to calibrate the CLIPScore confidence intervals described in the previous section, and assess both the human judgment correlation (Kendal-$\tau_C$), and the correlation between prediction errors and uncertainty estimates (UPS).
As mentioned in Section \ref{subsec:risk_control_ci}, to calibrate the scaling factor of the standard deviation, we use the Uncertainty Pearson Risk (UPR) function shown in Equation \ref{eq:UPR}. To evaluate our results, we use UPS and Accuracy.
In Appendix \ref{app:metrics}, we provide a more detailed explanation of each metric. In turn, Appendix \ref{app:datasets} further details the datasets. \looseness-1

\subsubsection{Guarantees on CLIPSCore Distributions}
In this section, we evaluate the performance gains achieved through the risk control calibration process applied to CLIPScore distributions obtained by fitting a truncated Gaussian to the output distributions of the attention sampling method. Our primary objective is to improve the correlation between prediction errors and uncertainty estimates (i.e., the standard deviation), which is measured by the UPS metric, while preserving overall system performance on external metrics, specifically by maintaining a strong correlation between the interval’s mean value and human judgments.

%\begin{figure}
%    \centering
%    \includegraphics[width=1\linewidth]{Figures/UPearsonS_UB.png}
%    \caption{Illustration of the calibration process using the model size B, and VICR validation set for calibration.}
%    \label{fig:ci_calibration}
%\end{figure}

%Figure \ref{fig:ci_calibration} illustrates the calibration process described in Section \ref{subsec:risk_control_ci}. This process involves selecting $\lambda^*$, corresponding to the lowest $p$-value below the risk tolerance.

\begin{table}[t!]
\centering
\renewcommand{\arraystretch}{1.3}
\resizebox{\columnwidth}{!}{%
\begin{tabular}{lcccccccccccc}
 & \multicolumn{2}{c}{\textbf{VICR}} & & \multicolumn{2}{c}{\textbf{Polaris}} && \multicolumn{2}{c}{\textbf{EX-8K}} && \multicolumn{2}{c}{\textbf{COM}} \\ \cline{2-3} \cline{5-6} \cline{8-9} \cline{11-12} 
\textbf{Method} & \textbf{UPS}    & \textbf{$\tau_c$}    && \textbf{UPS}     & \textbf{$\tau_c$}    && \textbf{UPS}     & \textbf{$\tau_c$}    &&  \textbf{UPS}    & \textbf{$\tau_c$}   \\ \hline
 \textbf{B-PRE}                   &  $22,1$         & $63,1$          &&      $38,1$      &  $50,1$         &&       $2,8$      &  $53,1$          &&   $18,3$        & $47,2$                \\  
 \textbf{B-POS}                   &  $36,4$         & $61,5$          &&      $44,1$      &  $49,4$         &&       $13,4$     &  $51,9$          &&   $26,1$        & $46,9$              \\ \cline{1-12}  
 \textbf{H-PRE}                   &  $42,6$         & $67,8$          &&      $60,2$      &  $51,0$         &&       $24,0$     &  $56,9$          &&   $18,3$        & $54,6$               \\  
 \textbf{H-POS}                   &  $49,6$         & $66,4$          &&      $70,1$      &  $50,6$         &&       $23,1$     &  $55,8$          &&   $27,1$        & $53,6$               \\ \hline
\end{tabular}
}
\vspace{-0.25cm}
\caption{Performance before (PRE) and after (POS) calibration of the CLIPScore confidence intervals across two model sizes: B (ViT-B/32) and H (ViT-H/14).}
\label{tab:ci_results}
\vspace{-0.25cm}
\end{table}

Table \ref{tab:ci_results} presents results before (PRE) and after (POS) calibration of the CLIPScore confidence intervals. For both model sizes, we achieve a significant improvement in performance in terms of UPS across all datasets, without significantly compromising the correlation with human ratings. Hence, our findings align with our original objective, providing a lightweight and model-agnostic methodology for obtaining more reliable confidence intervals over image captioning scores.

\section{Conclusions}

We proposed methods for producing and calibrating distributions on CLIPScore assessments, enabling granular caption evaluation and uncertainty representation. We leverage conformal risk control to address word-level foil detection and confidence estimation, allowing for flexible, task-specific risk-control with formal guarantees.
%demonstrating competitive performance against more complex models across well-established benchmarks while providing formal guarantees regarding risk control. 
The experimental results demonstrate competitive performance against more complex models on several well-established benchmarks. The proposed methods achieve  controllable and trustworthy performance in detecting foil words, and improved correlation between uncertainty estimates and prediction errors, without compromising human rating alignment. Our work highlights the potential of conformal calibration in enhancing the robustness and reliability of vision-and-language evaluation metrics.\looseness-1

\section*{Limitations and Ethical Considerations}

The research reported on this paper aims to enhance transparency and explainability, given that we advanced methods that can shed new light on the evaluation process of image captioning models. Specifically, our research introduced methods that offer uncertainty intervals and identify foil words within captions. It is important to note that our research does not specifically tackle potential biases in the CLIPScore evaluation metric (or biases existing in the popular benchmark datasets that also supported our experiments), nor does it address specific known limitations associated with CLIP models. Furthermore, our experiments were conducted exclusively in English, leaving open questions about the generalizability of our conformal risk control framework and word-level assessments to other languages, especially those with distinct morphological structures or additional syntactic complexities. Previous work has shown that uncertainty quantification methods are broadly applicable across languages, but often require language-specific calibration to ensure fair and balanced performance \cite{zerva-martins-2024-conformalizing}. Expanding our approach to linguistically diverse datasets is an important direction for future work.

Although our method improves interpretability and provides well-calibrated CLIPScore intervals, human evaluation remains indispensable to ensure reliable model assessments. Automated metrics should complement, not replace, human judgments, especially in sensitive applications, where misinterpretations can have significant consequences.

The method used to generate score distributions strongly depends on the length of the captions, which poses a limitation in our current framework. When captions are extremely short, such as a single word identifying the most prominent object, our approach struggles to produce meaningful uncertainty estimates, though CLIPScore distributions can still be generated with image-mask-based sampling. Addressing this limitation is an important direction for future research. Key priorities include enhancing linguistic diversity, improving uncertainty quantification, and integrating large-scale human validation to strengthen the overall robustness of the approach.

We also note that we used GitHub Copilot during the development of our research work, and we used ChatGPT for minor verifications during the preparation of this manuscript.

\section*{Acknowledgements}
We thank the anonymous reviewers for their valuable comments and suggestions. This research was supported by the Portuguese Recovery and Resilience Plan through project C645008882-00000055 (i.e., the Center For Responsible AI), by Fundação para a Ciência e Tecnologia (FCT) through the projects with references 2024.07385.IACDC and UIDB/50021/2020 (DOI:10.54499/UIDB/50021/2020), by EU's Horizon Europe Research and Innovation Actions (UTTER, contract 101070631), and also by FCT/MECI through national funds and, when applicable, co-funded EU initiatives under contract UID/50008 for Instituto de Telecomunicações.

\bibliography{custom}

\appendix
\section{The CLIPScore Metric} \label{app:clipscore}

We now formally describe the CLIPScore metric \cite{hesselclipscore}. In brief, CLIPScore is based on a modified cosine similarity between representations for the input image and the caption under evaluation. The image and the caption are both passed through the respective feature extractors from a given CLIP model. Then, we compute the cosine similarity of the resultant embeddings, adjusting the resulting value through a re-scaling operation. For an image with visual CLIP embedding $\textbf{v}$ and a candidate caption with textual CLIP embedding $\textbf{c}$, a re-scaling parameter is set as $w = 2.5$ and we compute the corresponding CLIPScore as follows:
\begin{equation}
\text{CLIPScore}({\textbf{c}}, {\textbf{v}}) =  w \times \max(\cos({\textbf{c}}, {\textbf{v}}), 0).
\end{equation}

Since CLIPScore is derived from a modified cosine similarity, it naturally inherits its bounded nature. As a result, CLIPScore values always fall within the interval $[0,2.5]$.

Note that the previous CLIPScore formulation does not depend on the availability of underlying references for each of the images in an evaluation dataset, hence corresponding to a reference-free image captioning evaluation metric.

\section{Proof of Proposition 1}
\label{appendix:proof_prop_1}

The proof for Proposition \ref{prop:ucb} uses the theorem of probability of subset events.
\begin{theorem} \label{theorem: subset_events}
    If $A$ and $B$ are events in a probability space such that $A \subseteq B$, then:
    \begin{equation}
        \mathbb{P}(A) \leq \mathbb{P}(B).
    \end{equation}
\end{theorem}
\noindent This is true because probability is additive over disjoint sets and satisfies:

\begin{equation}
    \mathbb{P}(B) = \mathbb{P}(A) + \mathbb{P}(B \textbackslash A),
\end{equation}
where $B \textbackslash A$ represents the part of $B$ not in $A$.

Using the previous theorem, the proof of Proposition \ref{prop:ucb} will be divided in three steps, which we describe next.

\textbf{Step 1.} Proof of the following equation:  
\begin{equation*}
    \mathbb{P}\Big(R(\lambda) > \hat{R}^+(\lambda) \Big) \leq \mathbb{P}\Big( g(\hat{R}(\lambda);R) < \delta \Big).
\end{equation*}

\noindent By construction, $R(\lambda) > \hat{R}^+(\lambda)$ implies that $g(R(\lambda);R) < \delta$, because $\hat{R}^+(\lambda)$ was chosen as the supremum of $R$ in the following set: 
\begin{equation*}
\Big\{R: g(\hat{R}(\lambda); R(\lambda)) \geq \delta \Big\}.
\end{equation*} 
This establishes that the event $R(\lambda) > \hat{R}^+(\lambda)$ necessarily leads to $g(R(\lambda);R) < \delta$. However, the converse does not hold. In other words, the event $R(\lambda) > \hat{R}^+(\lambda)$ is strictly contained within the event $g(R(\lambda);R) < \delta$. Applying Theorem~\ref{theorem: subset_events}, we can conclude that:
\begin{equation*}
    \mathbb{P}\Big(R(\lambda) > \hat{R}^+(\lambda) \Big) \leq \mathbb{P}\Big( g(\hat{R}(\lambda);R) < \delta \Big).
\end{equation*}

\noindent Next, let $G$ be the CDF of $\hat{R}(\lambda)$:
\begin{equation}
    G(t) = \mathbb{P}(\hat{R}(\lambda) \leq t).
\end{equation}
This implies that $G(t) \leq g(t;R(\lambda))$.

\textbf{Step 2.} Proof of the following equation:  
\begin{equation*}
    \mathbb{P}\Big(g(\hat{R}(\lambda);R) < \delta \Big) \leq \mathbb{P}\Big( G(\hat{R}(\lambda)) < \delta \Big).
\end{equation*}

\noindent By definition, $g(t;R)$ serves as an upper bound of $G(t)$. Therefore, the event $g(\hat{R}(\lambda);R) < \delta$, necessarily leads to $G(\hat{R}(\lambda)) < \delta$.  However, the converse does not hold. Applying Theorem~\ref{theorem: subset_events}, we can conclude that:
\begin{equation*}
    \mathbb{P}\Big(g(\hat{R}(\lambda);R) < \delta \Big) \leq \mathbb{P}\Big( G(\hat{R}(\lambda)) < \delta \Big).
\end{equation*}

\textbf{Step 3.} Proof of the following equation: 
\begin{equation*}
    \mathbb{P}\Big(G(\hat{R}(\lambda)) < \delta \Big) \leq \mathbb{P}\Big( \hat{R}(\lambda) < G^{-1}(\delta) \Big).
\end{equation*}

\noindent By definition, $G^{-1}(\lambda) = \text{sup } \Big\{ x: G(x) \leq \delta \Big\}$, which means that $G^{-1}(\lambda)$ is the highest value satisfying $G(x) \leq \delta$. Therefore, this will always imply $x \leq G^{-1}(\lambda)$. However, the converse is not always guaranteed. Because the event $G(\hat{R}(\lambda)) < \delta$ is strictly contained within the event $\hat{R}(\lambda) < G^{-1}(\delta)$, we can apply Theorem \ref{theorem: subset_events}, proving:
\begin{equation*}
    \mathbb{P}\Big(G(\hat{R}(\lambda)) < \delta \Big) \leq \mathbb{P}\Big( \hat{R}(\lambda) < G^{-1}(\delta) \Big).
\end{equation*}

\noindent Finally, since the event $\hat{R}(\lambda) < G^{-1}(\delta)$ is strictly contained in $\hat{R}(\lambda) \leq G^{-1}(\delta)$, by applying Theorem \ref{theorem: subset_events} we have: 
\begin{equation*}
    \mathbb{P}\Big(\hat{R}(\lambda) < G^{-1}(\delta) \Big) \leq \mathbb{P}\Big( \hat{R}(\lambda) \leq G^{-1}(\delta) \Big).
\end{equation*}

\noindent Next, using the definition of $G(x)$, we have that:
\begin{equation*}
    \mathbb{P}\Big(\hat{R}(\lambda) \leq G^{-1}(\delta) \Big) = G(G^{-1}(\delta)),
\end{equation*}
which by definition leads to $G(G^{-1}(\delta)) \leq \delta$.

Combining all the inequalities proved in each step, we have that:
\begin{equation}
    \mathbb{P}\Big(R(\lambda) > \hat{R}^+(\lambda) \Big) \leq \delta.
\end{equation}
Inverting the probability expression yields:
\begin{equation}
    \mathbb{P}\Big(R(\lambda) \leq \hat{R}^+(\lambda) \Big) \geq 1 - \delta,
\end{equation}
thus completing the proof.

\section{Concentration Inequalities} \label{app:concentrations}
Concentration inequalities provide probabilistic bounds on the deviation of a random variable from its expected value, playing a crucial role in statistical learning theory and probability analysis. This section presents key concentration inequalities, including Hoeffding’s and Bentkus’ inequalities.
\begin{proposition}[Hoeffding’s inequality, tighter version \cite{hoeffding}] \label{prop:hoeffding}

Suppose that $g(t;R)$ is a nondecreasing function in $t\in \mathbb{R}$ for every $R$. Then, for any $t<R(\lambda)$, we have that:
\begin{equation*}
    \mathbb{P}\Big( \hat{R}(\lambda) \leq t\Big)  \leq \text{exp}\{ -n \cdot f(t;R(\lambda))\},
\end{equation*}
where: 
\begin{equation*}
    f(t;R) = t\cdot\text{log}\left(\dfrac{t}{R}\right) + (1-t)\cdot\text{log}\left(\dfrac{1-t}{1-R}\right).
\end{equation*}
\end{proposition}

The weaker Hoeffding inequality is implied by Proposition \ref{prop:hoeffding}, noting that $f(t;R) \geq 2(R-t)^2$.

\begin{proposition}[Bentkus inequality \cite{bentkus}] \label{prop:bentkus}
    Supposing the loss is bounded above by one, we have that:
    \begin{equation*}
        \mathbb{P}\Big( \hat{R}(\lambda) \leq t\Big)  \leq e\mathbb{P}\Big( \text{Bi}(n, R(\lambda)) \leq \lceil nt \rceil \Big),
    \end{equation*}
    where $\text{Bi}(n, p)$ denotes a binomial random variable with sample size $n$ and success probability $p$.
\end{proposition}

\section{Details on Metrics} \label{app:metrics}
This section provides a detailed overview of the metrics used for calibrating the controlled variable and the evaluation metrics applied throughout the two different types of experiments.

\subsection{Metrics Used as Risks}
The following metrics were used to calibrate the threshold on the experiments regarding the detection of misaligned words in the caption.

\textbf{False Discovery Rate (FDR):} This metric is a statistical concept used to control the expected ratio of the number of False Positive classifications (FP) over the total number of positive classifications, including True Positives, (FP + TP). Mathematically, the False Discovery Rate (FDR) is defined as:
\begin{equation}
    \text{FDR} = \dfrac{\text{FP}}{\text{FP} + \text{TP}}.
\end{equation}

\textbf{False Positive Rate (FPR):} This metric is a statistical measure used to evaluate the proportion of actual negative instances that are incorrectly classified as positive by a model. It represents the likelihood of a false alarm, where the model predicts a positive outcome when the true outcome is negative. Mathematically, the False Positive Rate (FPR) is defined as follows:

\begin{equation}
    \text{FPR} = \dfrac{\text{FP}}{\text{FP} + \text{TN}},
\end{equation}
where FP denotes the number of False Positives, and TN represents the number of True Negatives.

\subsection{Evaluation Metrics}
The following metrics were applied throughout the experiments used to evaluate our methods.

\textbf{F1-Score:} The F1-score is a harmonic mean of precision and recall, providing a single metric that balances both measures. It is particularly useful in scenarios where class imbalance exists, as it considers both False Positives (FP) and False Negatives (FN). Mathematically, the F1-score is defined as:
\begin{equation}
    \text{F1} = 2 \cdot \dfrac{\text{Precision} \cdot \text{Recall}}{\text{Precision} + \text{Recall}}.
\end{equation}

In turn, Precision is defined as TP/(TP + FP), and Recall is defined as TP/(TP + FN). The F1-score ranges from 0 to 1, where a higher value indicates better model performance in terms of balancing precision and recall.

\textbf{Average Precision (AP):} The Average Precision (AP) metric is commonly used in information retrieval and classification tasks, particularly for evaluating models with imbalanced datasets. It summarizes the precision-recall curve through the weighted mean of precision achieved at each recall threshold, with the increase in recall serving as the weight. Mathematically, it is defined as:

\begin{equation}
    \text{AP} = \sum_{n} (R_n - R_{n-1}) \cdot P_n,
\end{equation}
where $P_n$ and $R_n$ are the precision and recall at the $n$-th threshold. Average Precision (AP) provides a single score that reflects the model's ability to correctly rank positive instances, with values closer to 1 indicating better performance.

\textbf{Location Accuracy (LA):} Localization Accuracy (LA) measures the fraction of samples where we can correctly identify a hallucinated object, among samples that are known to contain hallucinated objects. A sample receives an LA of 1 if at least one of the predicted hallucinated objects was correct, and an LA of 1 if the minimum matching score was a true hallucination.

\textbf{Uncertainty Pearson Score (UPS):} This metric is a statistical measure used to evaluate the correlation between the absolute error of predictions and their associated uncertainty estimates. It quantifies how well the model's uncertainty estimates align with the actual prediction errors, providing insights into the reliability of the uncertainty quantification. Mathematically, the Uncertainty Pearson Score (UPS) is defined as follows:

\begin{equation}
    \text{UPS} = \text{P}_C\left(|\mu(\lambda) - y|, \sigma(\lambda)\right),
\end{equation}
where \(|\mu(\lambda) - y|\) represents the absolute error between the predicted value \(\mu(\lambda)\) and the true value \(y\), and \(\sigma(\lambda)\) is the estimated uncertainty. A higher UPS indicates better calibration of uncertainty estimates, as it reflects a stronger correlation between prediction errors and uncertainty. 

\textbf{Kendall Tau C:} Seeing an evaluation dataset as a set of $n$ observations with the form $(\hat{y}_1,y_1), \ldots, (\hat{y}_n,y_n)$, for predicted scores $\hat{y}_i$ and reference ratings $y_i$, the Kendall Tau C correlation coefficient assesses the strength of the ranking association between the predicted scores and the reference ratings. Unlike Kendall Tau B, which accounts for ties, Kendall Tau C is specifically designed to handle cases where the underlying scales of the scores are different, such as when the number of possible ranks for the predicted scores and the reference ratings differ.

A pair of observations $(\hat{y}_{i},y_{i})$ and $(\hat{y}_{j},y_{j})$, where $i < j$, is considered concordant if the sort order of the instances agrees (i.e., if either both $\hat{y}_{i} > \hat{y}_{j}$ and $y_{i} > y_{j}$ hold, or both $\hat{y}_{i} < \hat{y}_{j}$ and $y_{i} < y_{j}$ hold). Otherwise, the pair is discordant. The Kendall Tau C coefficient is defined as:

\begin{equation}
\tau_{c}={\frac {n_{c}-n_{d}}{n_{0}}} \times {\frac {n-1}{n}} \times {\frac {m}{m-1}},
\end{equation}
where $n_{c}$ is the number of concordant pairs, $n_{d}$ is the number of discordant pairs, $n_{0} = n(n-1)/2$ is the total number of possible pairs, and $m$ is the number of distinct values in the ranking scale for the reference ratings. The term $\frac{m}{m-1}$ adjusts for the difference in scale between the predicted scores and the reference ratings, making Kendall Tau C particularly suitable for datasets that feature unequal ranking scales in the predictions and the ground-truth reference scores.

\section{Description of the Datasets} \label{app:datasets}
The following datasets were used in the calibration and evaluation of our method for detecting misaligned words in captions.
\begin{itemize}\setlength\itemsep{0em}
\item \textbf{Foil-it \cite{shekhar2017foil}:} The Foil-it dataset is a synthetic hallucination dataset based on samples from the MS-COCO \cite{lin2014microsoft} dataset. In this dataset, for each candidate-image pair, a “foil" caption is created which swaps one of the objects (in the MS-COCO detection set) in the caption with a different and closely related neighbour (chosen by hand to closely match, but aiming to be visually distinct). In our experiments, we used the test split of the Foil-it dataset, which includes $198,8814$ unique image-caption pairs. For calibration, we used $10\%$ of the validation split, which comprises a total of $395,300$ unique image-caption pairs.

\item \textbf{Foil-nocaps \cite{petryk2024aloha}:} The FOIL-nocaps dataset was introduced to address limitations of the FOIL-it dataset, which is overly biased towards object classes present in the MS-COCO dataset. The FOIL-nocaps dataset is based on the nocaps dataset \cite{agrawal2019nocaps}, which consists of images from the OpenImages dataset annotated with captions in a style similar to MS-COCO. The nocaps dataset is divided into three subsets (i.e., in-domain, near-domain, and out-of-domain) based on the relationship of the objects in the images to those in the MS-COCO dataset. Compared to Foil-it, this new dataset aims to provide a more general benchmark for evaluating hallucination detection methods, by including a broader range of object categories and contexts. In our tests, we used the test split of the Foil-nocaps dataset, which includes $5,000$ unique image-caption pairs. 

\item \textbf{Rich-HF \cite{liang2024rich}:} The Rich-HF dataset is a comprehensive benchmark for evaluating text-to-image alignment, comprising $18K$ image-text pairs with rich human feedback. It was constructed by selecting a diverse subset of machine generated photo-realistic images from the Pick-a-Pic \cite{kirstain2023pick} dataset, ensuring balanced use of categories such as ‘human’, ‘animal’, ‘object’, ‘indoor scene’, and ‘outdoor scene’. The dataset is annotated using the PaLI \cite{chenpali} visual question answering model, to extract basic features and ensure diversity. Rich-HF includes $16K$ training samples, $955$ validation samples, and $955$ test samples, with additional human feedback collected on unique prompts and their corresponding images. The dataset provides word-level misalignment annotations and overall alignment scores, making it a valuable resource for evaluating fine-grained text-to-image alignment and hallucination detection methods. Additionally, Rich-HF includes $955$ prompt-image pairs with detailed word-level misalignment annotations, covering a wide range of caption lengths, styles, and contents, due to its collection from real users. In our tests, we used the test split of the Rich-HF dataset and, for calibration, we used the validation split.
\end{itemize}

\begin{figure}
    \centering
    \includegraphics[width=1\linewidth]{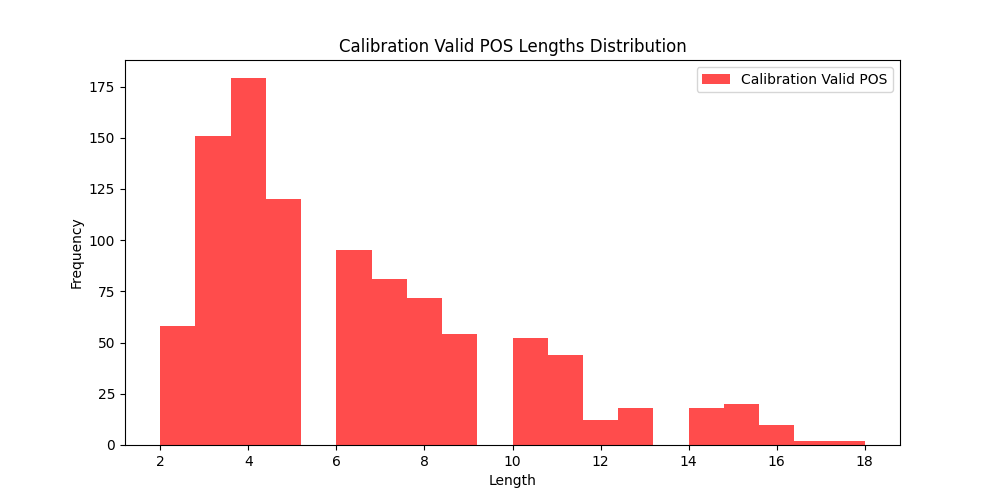}
    \caption{Frequency of sequences, with a given length, featuring words with valid parts of speech used for attention mask sampling in the Rich-HF calibration set.}
    \label{fig:rich_hf_cal}
\end{figure}

\begin{figure}
    \centering
    \includegraphics[width=1\linewidth]{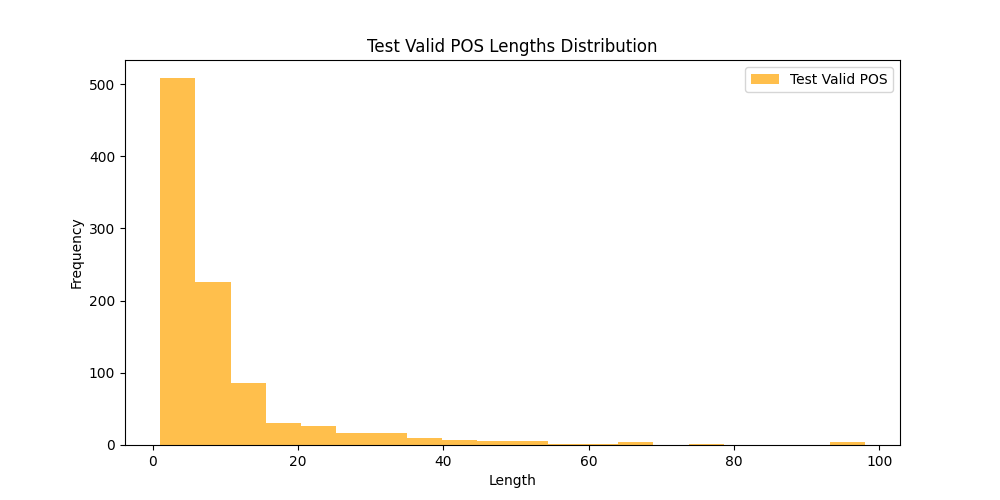}
    \caption{Frequency of sequences, with a given length, featuring words with valid parts of speech used for attention mask sampling in the Rich-HF test set.}
    \label{fig:rich_hf_test}
\end{figure}

While calibrating our methods using the Rich-HF dataset, we observed a significant difference in the distribution of the number of words per caption, between the calibration and test sets. Specifically, this disparity applies to words corresponding to valid parts of speech used in our attention mask sampling method, namely, nouns, proper nouns, numerals, verbs, adjectives, and adverbs. As noted in the main manuscript, this variation directly impacts the applicability of the thresholds. Figures \ref{fig:rich_hf_cal} and \ref{fig:rich_hf_test} show histograms illustrating the frequency of sequences, with a given length, featuring words with valid parts of speech used for attention mask sampling in the calibration and test sets, respectively. The figures illustrate the significant differences.

The following datasets were used in the calibration and evaluation experiments that assessed the Uncertainty Pearson Score (UPS), and correlation with human judgments.
\begin{itemize}\setlength\itemsep{0em}
\item \textbf{Flickr8K-Expert \cite{hodosh2013framing}:}
This dataset comprises $16,992$ expert human judgments for $5,664$ image-caption pairs from the Flickr8K dataset. Human assessors graded captions on a scale of $1$ to $4$, where $4$ indicates a caption that accurately describes the image without errors, and $1$ signifies a caption unrelated to the image.

\item \textbf{Composite \cite{aditya2015images}:} This dataset contains $13,146$ image-caption pairs taken from MS-COCO (2007 images), Flickr8K ($997$ images), and Flickr30K ($991$ images). Each image originally had five reference captions. One of these references was chosen for human rating and subsequently removed from the reference set that is to be used when assessing evaluation metrics.

\item \textbf{VICR \cite{narins2024validated}:} The Validated Image Caption Rating (VICR) dataset features 68,217 ratings, collected through a gamified approach, for $15,646$ image-caption pairs involving $9,990$ distinct images. The authors of the dataset demonstrated that it exhibits a superior inter-rater agreement compared to other alternatives (e.g., an improvement of $19\%$ in Fleiss’ $\kappa$ when compared to the agreement for the Flickr8K-Expert dataset), and it features a more balanced distribution across various levels of caption quality. In our tests, we used the test split of the VICR dataset, which includes $3,161$ unique image-caption pairs, with $2,000$ images from the MS-COCO 2014 validation dataset and $1,161$ images from the Flickr8K dataset. For calibration, we used the validation split, which comprises $2,310$ unique image-caption pairs.

\item \textbf{Polaris \cite{wada2024polos}:} The Polaris dataset comprises $131,020$ human judgments on image-caption pairs, collected from $550$ evaluators. It surpasses existing datasets in scale and diversity, offering an average of eight evaluations per caption, significantly more than Flickr8K (three) and CapEval1K (five). Polaris includes captions generated by ten standard image captioning models, covering both modern and older architectures to ensure output diversity. In our tests, we used the test split of the Polaris dataset, which includes $8,726$ unique image-caption pairs. For calibration, we used the validation split, which comprises $8,738$ unique image-caption pairs.
\end{itemize}

\section{Qualitative Results} \label{app:qualitative}

\begin{figure*}
    \centering
    \includegraphics[width=1\linewidth]{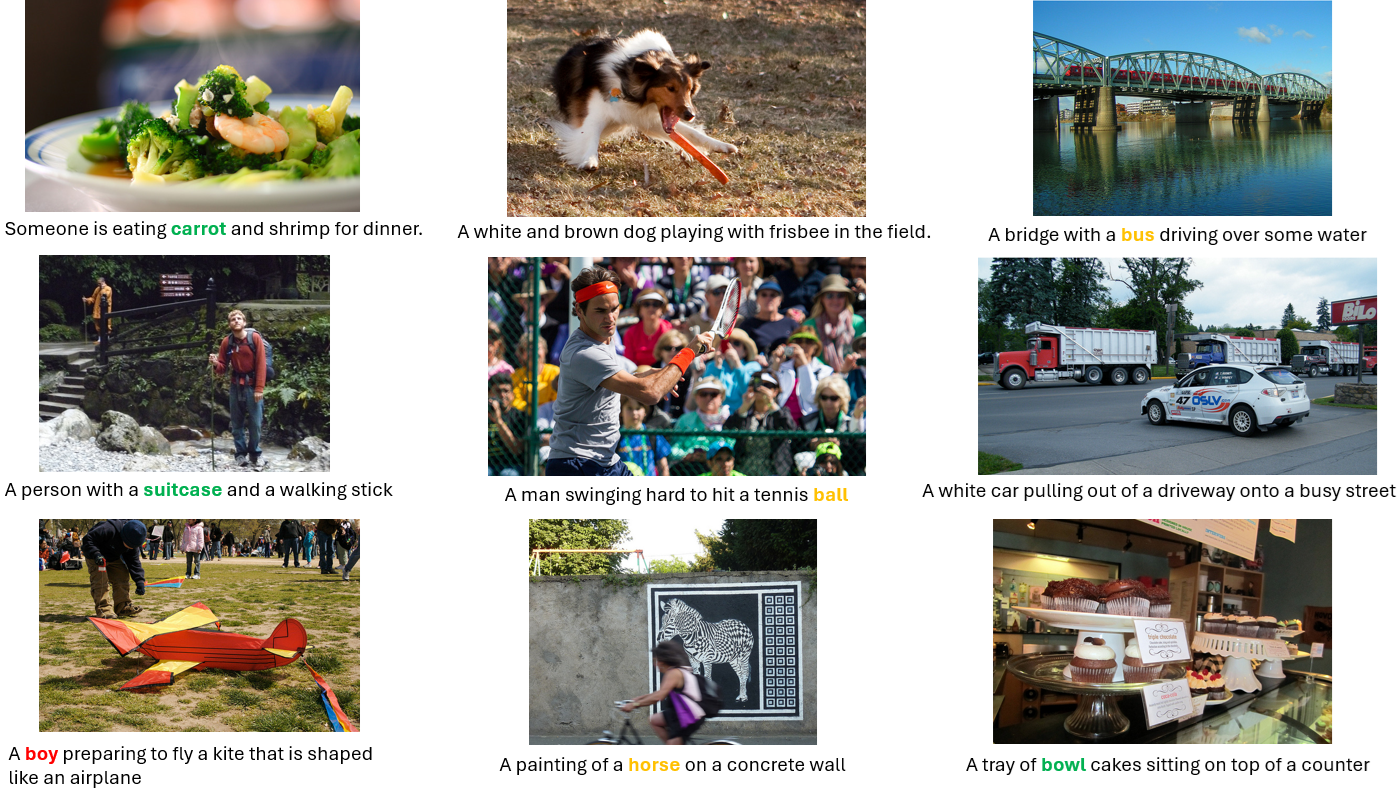}
    \caption{Qualitative results of the calibrated sampling method using the multilingual LAION ViT-H/14 CLIP model on the Foil-it test set. For these results, our method was calibrated to 20\% False Discovery Rate (FDR) using the Foil-it validation set for calibration.}
    \label{fig:foil-it_qualitative}
\end{figure*}

\begin{figure*}
    \centering
    \includegraphics[width=1\linewidth]{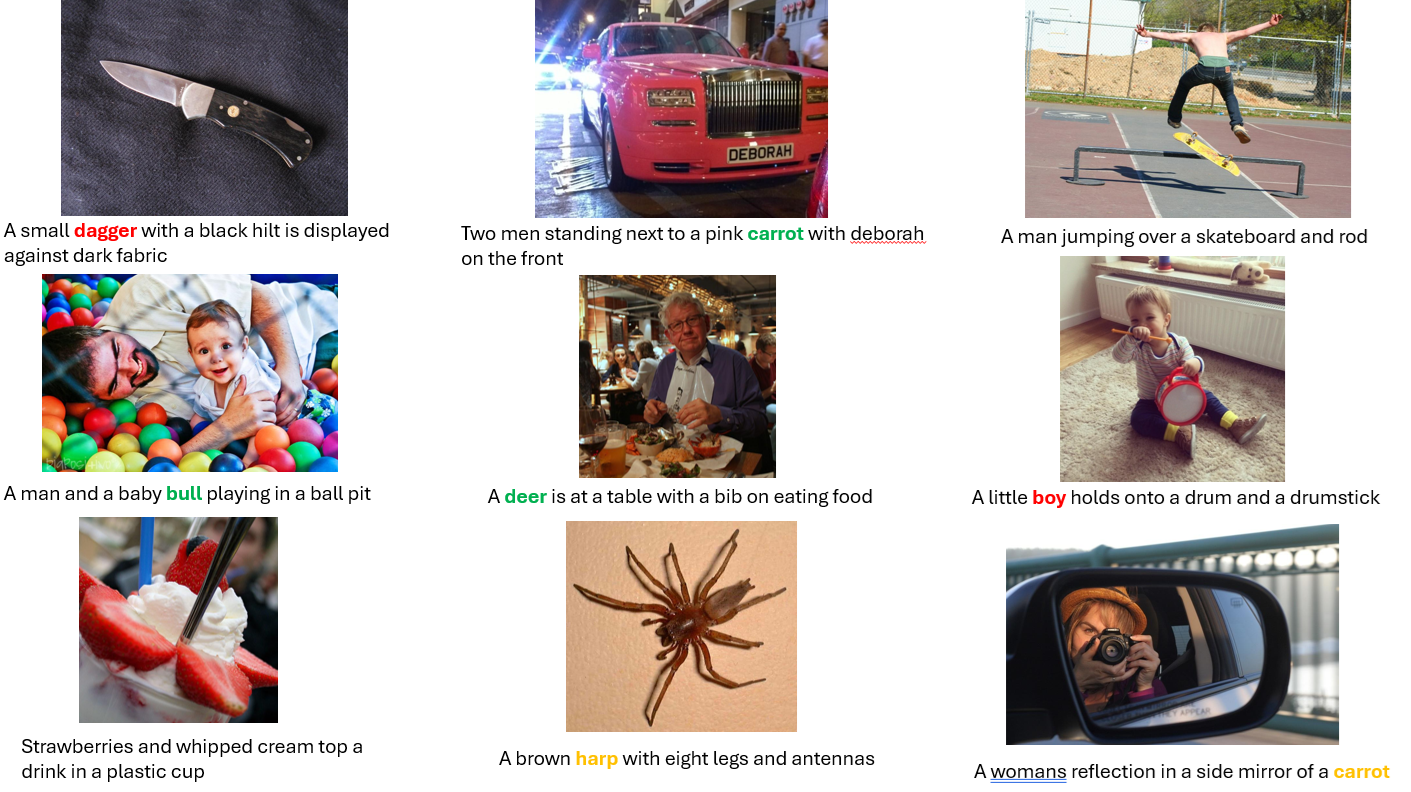}
    \caption{Qualitative results of the calibrated sampling method using the multilingual LAION ViT-H/14 CLIP model on the Foil-nocaps test set. For these results, our method was calibrated to 20\% False Discovery Rate (FDR) using the Foil-it validation set for calibration.}
    \label{fig:foil-nocaps_qualitative}
\end{figure*}

\begin{figure*}
    \centering
    \includegraphics[width=1\linewidth]{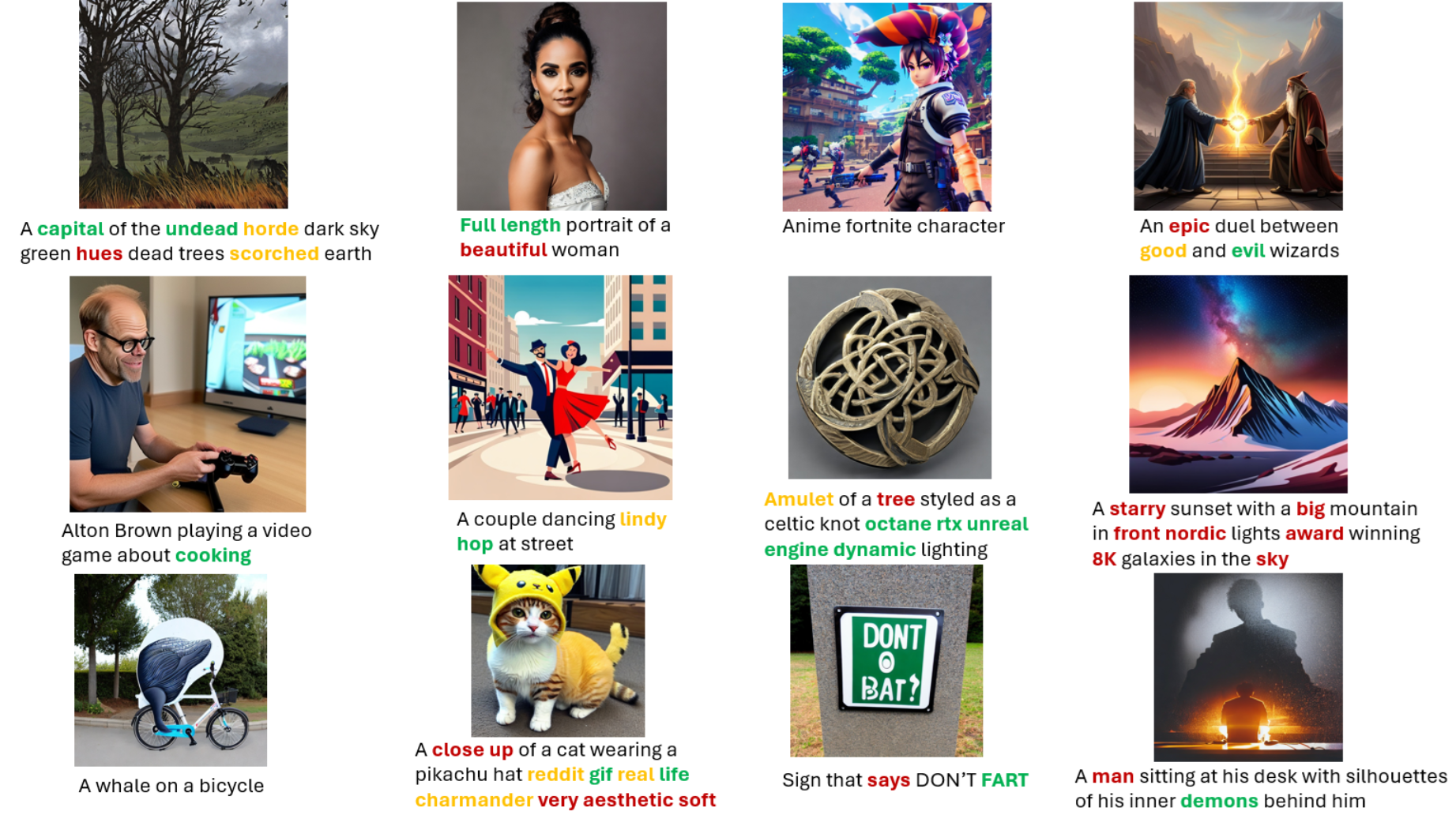}
    \caption{Qualitative results of the calibrated sampling method using the multilingual LAION ViT-H/14 CLIP model on the Rich-HF test set. For these results, our method was calibrated to 20\% False Discovery Rate (FDR) using the Rich-HF validation set for calibration.}
    \label{fig:rich_hf_qualitative}
\end{figure*}

We conducted a small qualitative study with the multi-class classification task of detecting misaligned words in the Foil-it (Figure \ref{fig:foil-it_qualitative}) and Foil-nocaps (Figure \ref{fig:foil-nocaps_qualitative}) benchmarks, as well as with the multi-label classification task using the Rich-HF benchmark (Figure \ref{fig:rich_hf_qualitative}). Throughout these qualitative experiments, captions associated with each image follow a color-coded scheme to indicate model performance in detecting misaligned words. Specifically, green highlights true positives, where our model correctly identified a misaligned word. Yellow indicates false negatives, meaning the model failed to detect an incorrect word. Lastly, red denotes false positives, where the model incorrectly flagged a word as misaligned when it was actually correct. Captions without coloured words are entirely correct according to the respective benchmark. This visual coding allows for an intuitive assessment of our model's strengths and weaknesses in the different benchmarks, across the examples.

\end{document}